\documentclass[conference]{IEEEtran}

\IEEEoverridecommandlockouts
\usepackage{cite}
\usepackage{amsmath,amssymb,amsfonts}
\usepackage{algorithmic}
\usepackage{graphicx}
\usepackage{textcomp}
\usepackage{xcolor}
\usepackage{subcaption} 
\usepackage{multirow}
\usepackage{tabularx}
\usepackage{amsmath}  
\usepackage{booktabs}

\usepackage{cite}

\def\BibTeX{{\rm B\kern-.05em{\sc i\kern-.025em b}\kern-.08em
    T\kern-.1667em\lower.7ex\hbox{E}\kern-.125emX}}
\usepackage{amsmath}
\begin{document}

\title{Frailty-Aware Transformer for Recurrent Survival Modeling of Driver Retention in Ride-Hailing Platforms

}

\author{\IEEEauthorblockN{Shuoyan Xu}
\IEEEauthorblockA{\textit{Civil \& Mineral Engineering} \\
\textit{University of Toronto}\\
Toronto, Canada \\
shuoyan.xu@mail.utoronto.ca}
\and
\IEEEauthorblockN{Yu Zhang}
\IEEEauthorblockA{\textit{Civil \& Mineral Engineering} \\
\textit{University of Toronto}\\
Toronto, Canada \\
yyu.zhang@utoronto.ca}
\and
\IEEEauthorblockN{Eric. J. Miller}
\IEEEauthorblockA{\textit{Civil \& Mineral Engineering} \\
\textit{University of Toronto}\\
Toronto, Canada \\
eric.miller@utoronto.ca}

}

\maketitle

\begin{abstract}
 Ride-hailing platforms are characterized by high-frequency, behavior-driven environments, such as shared mobility platforms. Although survival analysis has been widely applied to recurrent events in other domains, its use for modeling ride-hailing driver behavior remains largely unexplored. To the best of our knowledge, this study is the first to formulate driver idle behavior as a recurrent survival process using large-scale platform data. This study proposes a survival analysis framework that uses a Transformer-based temporal encoder with causal masking to capture long-term temporal dependencies and embeds driver-specific embeddings to represent latent individual characteristics, significantly enhancing the personalized prediction of driver retention risk, modeling how historical idle sequences influence the current risk of leaving the platform via trip acceptance or log-off. The model is validated on datasets from the City of Toronto over the period January 2 to March 13, 2020. The results show that the proposed Frailty-Aware Cox Transformer (\emph{FACT}) delivers the highest time-dependent C-indices and the lowest Brier Scores across early, median, and late follow-up, demonstrating its robustness in capturing evolving risk over a driver's lifecycle. This study enables operators to optimize retention strategies and helps policy makers assess shared mobility’s role in equitable and integrated transportation systems. 

\end{abstract}

\begin{IEEEkeywords}
survival analysis, driver retention, transformer, ride-hailing, recurrent events
\end{IEEEkeywords}

\section{Introduction}
The purpose of this study is to model the driver retention behavior through a transformer-based survival model. Shared mobility services, such as ride-hailing, car-sharing, and bike-sharing, are becoming an increasingly prominent component of contemporary transportation systems. These services are central to the broader concept of Mobility as a Service (MaaS) \cite{calderon2020literature}, which aims to integrate various forms of transport into a unified and user-centric platform. As shared mobility grows in scale and significance, its operational efficiency relies heavily on the ability to match travel demand with an adequate and responsive supply of drivers. A core challenge lies in understanding the behavior of both users and service providers. 

Driver behavior has key implications for system reliability. In some regions, platforms have experienced substantial driver attrition, often due to insufficient earnings or inefficient order assignment algorithms. In shared mobility systems, drivers exhibit flexible, self-determined work patterns that differ markedly from those in traditional fixed-schedule occupations. Drivers may repeatedly log in and out of the platform within a short period, resulting in a sequence of work shifts that are highly individualized and dynamic. This recurrent behavior introduces complex time-varying risks that cannot be adequately captured by conventional survival analysis methods, which typically assume a single event per subject. 

To the best of our knowledge, this study is the first to formulate driver idle and log-off behavior as a recurrent survival process using large-scale operational data. We introduce a transformer-based survival model that incorporates causal masking to capture long-term temporal dependencies in driver activity sequences, together with driver-specific embeddings to represent latent heterogeneity. This framework enables continuous estimation of a driver’s risk of logging off by leveraging the full historical record of idle and trip events. Beyond methodological novelty, the approach provides actionable insights for platform optimization and incentive design, and offers a foundation for integrating realistic supply-side dynamics into large-scale urban simulation models. 

This paper makes the following contributions:
\begin{enumerate}
    \item It explicitly models drivers' recurrent idle events, which effectively captures the dynamic and adaptive nature of driver participation in shared mobility platforms.
    \item     This study focuses on the application of a transformer-based encoder with causal masking to capture long-term temporal dependencies. Specifically, we focus on representing latent individual characteristics via the integration of driver-specific embeddings. This integration significantly enhances our ability to personalize the prediction of driver retention risk. 
  
\end{enumerate}

\section{Literature Review}

\subsection{Driver behavior and retention on ride-hailing platforms }

Recent literature identifies several factors directly influencing ride-hailing drivers’ decisions to end their shifts. Ashkrof did a PhD thesis on the supply side behavioral dynamics of ride-hailing platforms, and identified that drivers have an expected cumulative earning threshold, beyond which the continued working yields diminishing returns. It also concluded that operational inefficiencies like high idle times or long pickup distances, and spatial strategies where drivers strategically log off if positioned unfavourably or outside high-demand zones, are the main contributors \cite{ashkrof2023supply}. Ashkrof et al. further emphasize ride characteristics such as excessively long pickup times, low fares, and insufficient incentives (e.g., lack of surge pricing) are the key determinants in influencing drivers’ decisions to discontinue working; these decisions also vary significantly with drivers’ employment status, experience levels, and weekly scheduling preferences \cite{ashkrof2022ride}. Similarly, Hall and Krueger emphasized the role of flexibility and income smoothing, noting that drivers frequently end their shifts upon achieving personal earnings goals or daily financial targets \cite{hall2018analysis}. Ma et al. found that competitive market dynamics also lead drivers to terminate their shifts early when faced with high competition and reduced profitability \cite{ma2019modeling}. Qin et al. illustrate that inefficient dispatch algorithms causing dissatisfaction and prolonged idle periods strongly motivate drivers to prematurely sign off \cite{qin2020ride}. In sum, these studies indicate that cumulative earning and personal economic goals, idling time, order location, market competition, and algorithmic effectiveness (pickup time and distance) are the major determinants in ride-hailing drivers’ shift-ending behaviors.

Several indirect and less quantifiable factors also influence ride-hailing drivers’ decisions to end their shifts, typically involving psychological perceptions, subjective experiences, and individual circumstances. For instance, Chakraborty and Heeks note that perceived fairness and algorithmic transparency significantly impact drivers’ decisions; unmet earnings expectations and perceived unfair management practices often result in earlier shift-ends \cite{chakraborty2023gig}. Similarly, Sun et al. note that the dispatching method, i.e. inform vs. assign system, affects shift duration \cite{sun2020taxi}. Lefcoe et al. emphasize that drivers facing economic uncertainty often continue working despite fatigue or safety risks, while those who feel financially secure would sign off earlier when feeling tired or unsafe \cite{lefcoe2024ride}. Doğan et al. analyzed from the marketing perspective and suggested that psychological factors, such as satisfaction derived from clear goal-oriented feedback and performance-based incentives, substantially influence drivers’ motivation and indirectly reduce premature sign-offs \cite{dougan2024platform}. In addition, Berliner and Tal noted that driver attitudes, vehicle ownership, and life circumstances are important determinants on driver’s shift-ending decisions, due to personal responsibilities or lower perceived job value \cite{berliner2018drives}. These studies indicate that individual perception, satisfaction, experience and demographic attributes of the drivers, though challenging to measure to get data, are also important factors. This study would not consider these indirect and less quantifiable factors in the analysis. 

Apart from the research on contributing factors, some other interesting studies provide complementary insights into the broader dynamics of ride-hailing drivers’ decisions beyond immediate shift-end behaviors. For instance, Young et al. investigated the impact of regulatory pressures and changing cost structures on driver behaviors in the City of Toronto, and found that increased operational complexity in licensing, insurance, rising vehicle-related expenses can indirectly lead drivers to permanently leave the platform \cite{young2024road}. Hegde et al. investigated ride-hailing driver attrition during COVID-19 \cite{hegde2024characterizing}. The findings include that drivers logged off permanently or temporarily to avoid infection risks; lower demand led to increased idle times and lower efficiency; and that drivers using multiple platforms were less likely to log off because they could secure higher trip frequency and income. Zhou et al. explored the psychological and organizational dimensions, and found that salary dissatisfaction, inadequate work-life balance, poor organizational support, and high perceived workloads are significant predictors of drivers’ turnover intentions \cite{zhou2024drives}. Although these factors may not directly determine daily shift decisions, they substantially influence broader career choices and overall driver availability, which influence the long-term dynamics of platform labor supply.

\subsection{Survival analysis, recurrent event and ride-hailing}

Survival analysis is a statistical technique used to analyze the expected duration of time until one or multiple events occur, often handling data characterized by censoring—instances where the event of interest has not yet happened by the observation time or the participant’s withdrawal \cite{fox2002cox,kelly2000survival, kleinbaum2012kaplan}. Traditional survival analysis focuses on single-event outcomes, such as death or failure, but in many practical scenarios, especially in clinical and epidemiological studies, events occur repeatedly over time \cite{cook2007statistical}. To handle these recurrent events appropriately, several extended models have been developed. These include the Andersen-Gill (AG) model \cite{andersen2012statistical,andersen1982cox}, which treats recurrent events as counting processes, the Prentice-Williams-Peterson (PWP) models \cite{prentice1981regression}, and frailty models \cite{duchateau2008frailty}, which incorporate random effects to account for unobserved individual variability \cite{cook2007statistical,twisk2005applied}. Recurrent-event survival analysis models have been widely applied to cases like repeated hospitalizations or infections, and repeated system failures, which demonstrate their versatility in modeling correlated and repetitive phenomena over time \cite{kostic2021deep,thenmozhi2019survival}.

Specifically among the survival model developed and applied to recurrent event, frailty model captures the unobserved heterogeneity among the samples with a random factor (i.e., the ``frail") \cite{liu2004shared}. In recurrent events, not all samples with the same covirates carry the same risk over time, and the "frail" allows for shared information across repeated events within a subject and account for correlation in recurrent event data \cite{amorim2015modelling}. Frailty model is widely used in medical research to study recurrent event processes such as hospital readmissions and infectious disease episodes \cite{rogers2016analysis, tawiah2019frailty, huang2007joint}. It has also been applied in industrial reliability and insurance, where systems or policies experience repeated failures or claims \cite{pena2004models, brown2023reliability}.

Despite the widespread application of survival analysis to recurrent events, its utilization in analyzing driver behavior in ride-hailing contexts remains limited. Although survival models have been effectively employed in transportation and shared mobility studies, such as predicting vehicle idle times or car-sharing availability \cite{kostic2021deep}, specific applications to individual ride-hailing driver decisions, such as shift-end behaviors or platform log-off decisions, are rarely applied in existing literature. Studies typically focus on aggregate fleet management or demand forecasting, rather than individual level recurrent event modeling of driver behaviors \cite{fathi2024ride}. Given that ride-hailing drivers experiences recurrent decisions to make whether or not to end the shift after recurrent order completion, it is promising to develop a survival model to improve predictions of driver retention and activity patterns within ride-hailing platforms. 

\subsection{Deep Learning for Survival Analysis}

The availability of large-scale behavioral and temporal data has attracted growing interest in applying deep learning to survival analysis, primarily due to its ability to model complex nonlinear relationships~\cite{kvamme2019time, katzman2018deepsurv, faraggi1995neural,lin2022deep}, relax restrictive statistical assumptions~\cite{katzman2018deepsurv, lee2018deephit, lee2019dynamic,ren2019deep}, and generate personalized predictions~\cite{katzman2018deepsurv, lee2018deephit}. Within this context, Katzman \emph{et al.} proposed DeepSurv~\cite{katzman2018deepsurv}, a multilayer perceptron (MLP) model designed to capture nonlinear dependencies within the Cox proportional hazards framework. Inspired by recommender systems, DeepSurv generates personalized treatment recommendations by learning individual risk representations, allowing the baseline hazard to vary across different patient subgroups. To address time-varying covariates, Cox-Time extends the Cox model by allowing time-dependent effects of features on the hazard function. Beyond the proportional hazards assumption, DeepHit~\cite{lee2018deephit} relaxes the underlying assumptions by discretizing time into intervals and directly estimating the joint distribution of event type and time interval. To handle right-censored data, it employs a combination of ranking loss and negative log-likelihood loss. 

For survival prediction involving longitudinal inputs, sequence models such as RNNs~\cite{grossberg2013recurrent}, LSTMs~\cite{hochreiter1997long}, and Transformers~\cite{vaswani2017attention,lin2022deep} have been adopted to capture temporal dependencies. For instance, Dynamic-DeepHit~\cite{lee2019dynamic} further extends DeepHit by integrating a recurrent neural network with attention mechanisms to process longitudinal covariates and dynamically update survival predictions over time. It supports competing risks and outputs the joint distribution of event type and event time, enabling personalized, real-time risk assessment in sequential clinical settings. Ren \emph{et al.}~\cite{ren2019deep} applies the probability chain rule to derive both the survival function and event time distribution, while incorporating losses for censored and uncensored samples.  While these models have demonstrated effectiveness in single-event survival prediction, they do not naturally extend to recurrent event settings where the same individual may experience repeated instances of the same event~\cite{cai2003analysis}. More specifically, they lack mechanisms to model inter-event dependencies. Our proposed model focuses on recurrent survival prediction, where each driver session is treated as a distinct event. We propose a Transformer-based temporal encoder combined with learnable vehicle embeddings to model dependencies across multiple driver sessions. This design enables the model to capture long-term behavioral patterns and personalize survival predictions in high-frequency, behavior-driven environments such as ride-hailing platforms.

\section{Methodology}\label{AA}

\subsection{Problem Definition}
\begin{figure}[htbp]
  \centering
  \includegraphics[width=\columnwidth]{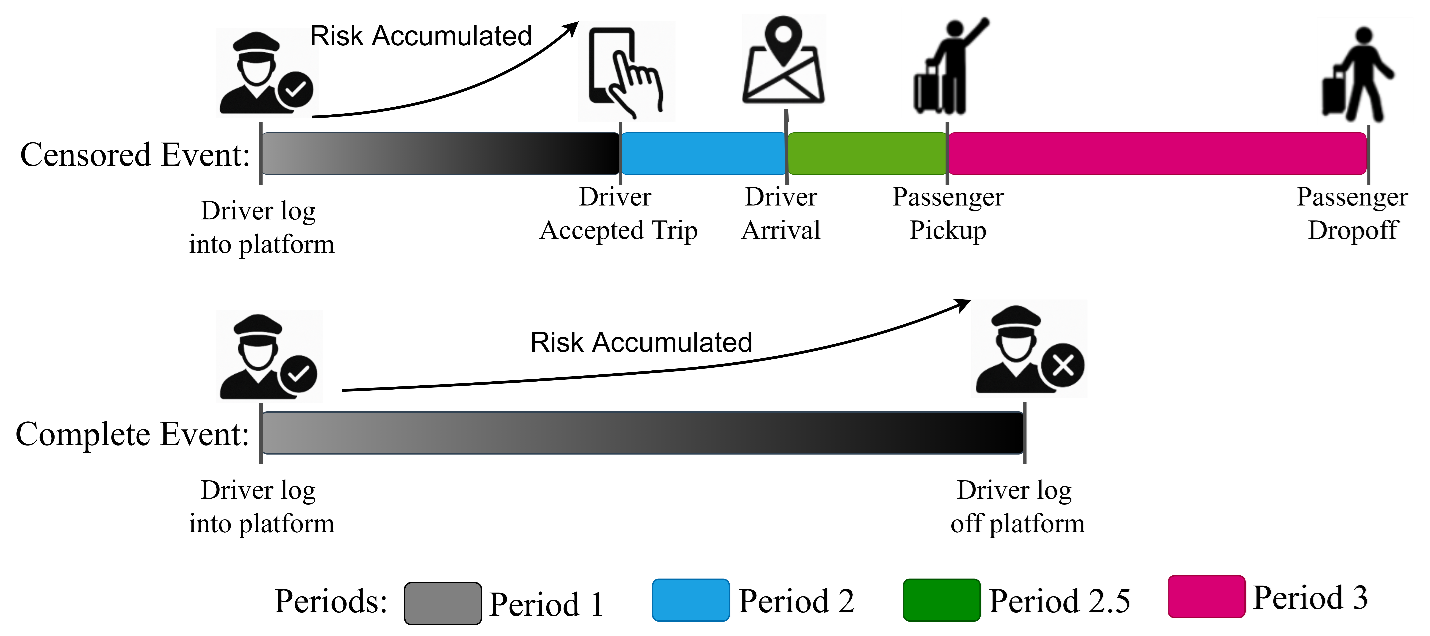}
  \caption{Illustration of censored and complete events defined in ride-hailing}
  \label{fig:idle-events}
\end{figure}

The ride-hailing driver retention process can be conceptualized as a repeated sequence of idle periods initiated by drivers, each ending either with a trip request or a decision to log off the platform. As shown in Fig.~\ref{fig:idle-events}, In this study, each idle interval begins with a “birth” event and ends with a “death” event or a “censored” event. A “death” event corresponds to the natural driver log-off behavior, while a “censored” event refers to an idle state interrupted by a trip, making the true end of the interval unobservable. This study employs survival analysis to estimate the instantaneous hazard of leaving the platform. In our formulation, hazard accumulation is defined to occur only during idle intervals, as drivers cannot exit the platform while actively servicing a trip. Accordingly, we frame each driver's idle period as a time-to-event problem, where the origin is the start of the idle period and the event of interest is its termination. The objective of the study is to estimate the hazard function:
\begin{equation}
\
h(t \mid X) \;=\; \lim_{\Delta t \to 0} \frac{\Pr\bigl(t \le T < t + \Delta t \mid T \ge t,\,X\bigr)}{\Delta t},
\
\end{equation}
which quantifies the instantaneous risk of idle-period termination given covariates \(X\).  

\subsection{Frailty-aware Transformer-Based Risk Function}

\begin{figure*}[htbp]
  \centering
  \includegraphics[width=\textwidth]{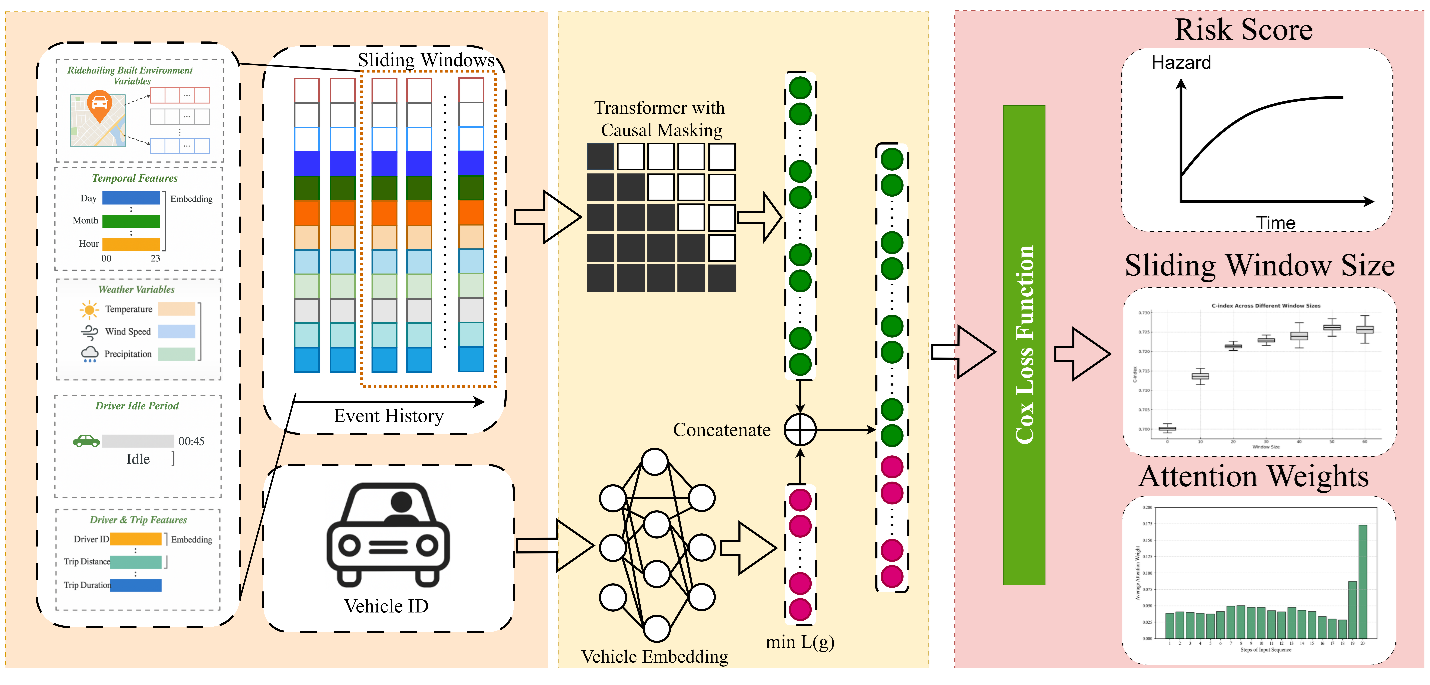}
  \caption{Framework of the proposed model}
  \label{fig:two-column-model}
\end{figure*}

In shared mobility platforms, drivers autonomously determine when to log on and off, leading to repeated idle periods throughout their engagement. To model this behavior, we formulate a recurrent event survival analysis framework that captures both the voluntary and temporal dynamics of driver participation. Each driver may experience multiple idle episodes over time. Let:
\begin{itemize}
    \item \(N\) be the total number of drivers, indexed by \(i \in \{1,2,\ldots,N\}\);
    \item \(K_i\) be the number of idle events experienced by driver \(i\), indexed by \(k \in \{1,2,\ldots,K_i\}\).
\end{itemize}

For each idle event \((i,k)\), we define:
\begin{itemize}
    \item \(X_{i,k} \in \mathbb{R}^p\): covariate vector at the start of the idle period (e.g., cumulative earnings, order counts, travel distances, and temporal encodings);
    \item \(T_{i,k}\): the observed duration of the idle event;
    \item \(\delta_{i,k} \in \{0,1\}\): the event indicator, defined as
    \begin{equation}
        \delta_{i,k} =
        \begin{cases}
            1, & \text{if the idle period ends naturally;} \\
            0, & \text{if the event is censored.}
        \end{cases}
        \label{eq:event_indicator}
    \end{equation}
\end{itemize}

To model temporal dependencies, we adopt a look back window approach. For each driver, we construct the input sequence using previous \(W\) consecutive events:
\begin{equation}
\{[X_{i,1}, y_{1}],\; [X_{i,2}, y_{2}],\; \ldots,\; [X_{i,W-1}, y_{W-1}],\; [X_{i,W}, \mathbf{0}]\},
\end{equation}
where \(y_t = [T_{i,t}, \delta_{i,t}]\) denotes the observed outcome of event \(t\), and \(\mathbf{0}\) is a zero placeholder for the target event. This formulation allows the model to learn from historical outcomes while predicting the current event.

We define the hazard function for event \((i,k)\) at elapsed time \(t\) as:
\begin{equation}
h_{i,k}(t \mid X_{i,1:k}, y_{1:k-1}, e_i; \theta) =
h_0(t) \exp\left( 
r\big(X_{i,1:k}, y_{1:k-1}, e_i; \theta\big)
\right),
\label{eq:hazard_function}
\end{equation}
where:
\begin{itemize}
    \item \(h_0(t)\): a shared baseline hazard function;
    \item \(X_{i,1:k}\): covariate sequence up to the \(k\)th event;
    \item \(y_{1:k-1}\): historical outcomes for the first \(k{-}1\) events;
    \item \(e_i \in \mathbb{R}^q\): a learnable, driver-specific frailty embedding;
    \item \(r(\cdot)\): a dynamic risk function parameterized by \(\theta\).
\end{itemize}

To capture individual heterogeneity, this study introduces a frailty embedding \(e_i\) into the risk function. The embedding encodes latent driver-specific factors, such as behavioral tendencies, that are not directly observed in the covariates. It is learned jointly with model parameters and remains fixed across events for each driver. In addition to adjusting baseline risk, incorporating \(e_i\) allows interaction with dynamic input features, enabling more personalized and context-aware hazard estimation.

The risk function \(r(\cdot)\) is modeled using a Transformer encoder equipped with causal self-attention, which enables the model to weigh the influence of each historical event. The concatenated covariate-outcome pairs are first projected into a latent space:
\begin{equation}
\mathbf{E}_t = \mathbf{W}_x [X_{i,t}, y_t] + \mathbf{b}_x, \quad t = 1, \ldots, W,
\end{equation}
where \(\mathbf{W}_x\) and \(\mathbf{b}_x\) are learnable parameters.

To retain temporal ordering, positional encodings are added to each \(\mathbf{E}_t\). The enriched embeddings are then passed through multiple layers of self-attention and feed-forward transformations:
\begin{equation}
\text{Attention}(Q, K, V) = \operatorname{softmax}\left(\frac{QK^\top}{\sqrt{d_k}}\right) V,
\end{equation}
where \(Q\), \(K\), and \(V\) are query, key, and value matrices derived from the input sequence, and \(d_k\) is the key dimension. Causal masking ensures that the model only attends to past and current time steps:
\begin{equation}
\operatorname{Mask}(Q,K)_{t,\tau} =
\begin{cases}
-\infty, & \tau > t, \\
0,       & \tau \leq t.
\end{cases}
\end{equation}

From the final layer, we extract the hidden representation corresponding to the most recent event:
\begin{equation}
z_{i,k} = \text{TransformerEncoder}(\{[X_{i,t}, y_t]\}_{t=1}^{W})_W.
\end{equation}
The final risk score is computed by fusing this representation with the driver-specific embedding:
\begin{equation}
r_{i,k} = f(z_{i,k}, e_i; \phi),
\end{equation}
where \(f(\cdot)\) is a learnable projection function parameterized by \(\phi\). This architecture enables the model to jointly capture both temporal dynamics and unobserved driver heterogeneity in estimating the risk of idle event termination.

\subsection{ Loss Function for Recurrent Events}
In this study, we use a recurrent event Cox formulation to model within-subject correlation and repeated idle episodes during a driver’s platform engagement. This approach addresses the limitations of the standard Cox model, which assumes independent observations and is limited to time-to-first-event data. Additionally, it accommodates right-censored data by excluding censored observations from the event contribution while retaining them in the risk set. This ensures that censored observations inform the likelihood calculation without biasing the hazard estimation. 

The partial likelihood for an individual event \((i,k)\) is formulated as:
\begin{equation}
\
L_{i,k} = \left[ \frac{\exp(r_{i,k})}{\sum_{(j,s) \in R(T_{i,k})} \exp(r_{j,s})} \right]^{\delta_{i,k}},
\
\end{equation}
where the risk set \( R(T_{i,k}) \) is defined as:
\begin{equation}
\
R(T_{i,k}) = \{ (j,s) : T_{j,s} \geq T_{i,k} \}.
\
\end{equation}
This set includes all events across all drivers that have a duration at least as long as \( T_{i,k} \), ensuring that only events at risk of occurring at or after \(T_{i,k}\) are considered. The partial likelihood across all drivers and their respective events is:
\begin{equation}
\mathcal{PL}(\theta, \phi, E) = \prod_{i=1}^{N} \prod_{k=1}^{K_i} 
\left[
\frac{\exp(r_{i,k})}{\sum\limits_{(j,s) \in R(T_{i,k})} \exp(r_{j,s})}
\right]^{\delta_{i,k}},
\end{equation}
To train the model, we minimize the negative log-partial likelihood, which defines our loss function as:

\begin{multline}
\mathcal{L}(\theta,\phi,E) 
= - \sum_{i=1}^{N}\sum_{k=1}^{K_i} \delta_{i,k} \Biggl[
     r_{i,k} \\
     - \log \Biggl(
         \sum_{(j,s)\in\mathcal{R}(T_{i,k})} \exp\!\bigl(r_{j,s}\bigr)
       \Biggr)
   \Biggr].
\end{multline}

\subsection{Evaluation Metrics}

To assess the performance of our survival analysis model, we employ two widely-used metrics: the concordance index (C-index) and the Brier score. The C-index measures the model's ability to correctly rank the risk of events, while the Brier score evaluates the accuracy and calibration of the predicted survival probabilities.

\subsubsection{Concordance Index (C-index)}
Let \((T_i, \delta_i)\) denote the observed survival time and event indicator for the \(i\)-th sample, and let \(r_i\) be the risk score predicted by the model. The concordance index is defined as the proportion of all usable pairs for which the ordering of the predicted risk scores is concordant with the ordering of the observed survival times:
\begin{equation}
C = \frac{\displaystyle \sum_{i,j} \mathbb{I}\{T_i < T_j\} \, \mathbb{I}\{r_i > r_j\} \, \delta_i}{\displaystyle \sum_{i,j} \mathbb{I}\{T_i < T_j\} \, \delta_i},
\label{eq:c_index}
\end{equation}

where \(\mathbb{I}\{\cdot\}\) denotes the indicator function, and the summation is taken over all pairs \((i,j)\) such that \(T_i < T_j\) and the event for subject \(i\) is observed (\(\delta_i=1\)). A C-index of 1 indicates perfect prediction, whereas a C-index of 0.5 implies random predictions.

\subsubsection{Brier Score}
In survival analysis, the Brier score at a given time \(t\) quantifies the mean squared error between the observed event status and the predicted survival probability. It is defined as:

\begin{equation}
\text{BS}(t) = \frac{1}{N} \sum_{i=1}^{N} w_i(t) \left( \mathbb{I}\{T_i \leq t,\, \delta_i=1\} - \hat{S}(t \mid x_i) \right)^2
\label{eq:brier_score}
\end{equation}

where:
\begin{itemize}
    \item \(N\) is the total number of subjects,
    \item \(\hat{S}(t \mid x_i)\) is the predicted survival probability for subject \(i\) at time \(t\),
    \item \(w_i(t)\) is a weight function to account for censoring, and
    \item \(\mathbb{I}\{T_i \leq t,\, \delta_i=1\}\) equals 1 if subject \(i\) has experienced the event by time \(t\) (and is uncensored) and 0 otherwise.
\end{itemize}

To summarize the performance over a time interval \([0,\tau]\), the Integrated Brier Score (IBS)  is computed as:

\begin{equation}
\text{IBS} = \frac{1}{\tau} \int_{0}^{\tau} \text{BS}(t) \, dt.
\label{eq:ibs}
\end{equation}

A lower Brier score (or IBS) indicates better predictive accuracy and calibration of the model. In our experiments, both the C-index and the Brier score (or IBS) are computed to evaluate both discriminative (ranking) performance and the overall predictive accuracy of the proposed model.

\section{Data}
\subsection{Data Source}
This study integrates four complementary datasets, including ride-hailing records, workshift metrics, weather, and spatiotemporal variables, to analyze driver retention on ride-hailing platforms.
\begin{table*}[htbp]
  \centering
  \caption{Variables and Descriptions}
  \label{tab:variables}
  \small
  \begin{tabular}{|c|c|p{11cm}|}
    \hline
    \textbf{Feature Type} & \textbf{Variable Name} & \textbf{Description} \\
    \hline
    \multirow{6}{*}{Temporal Features} 
        & timestamp        & Date and time when the record was captured. \\
        & hour\_sine       & Sine transformation of the hour to capture cyclical daily patterns. \\
        & hour\_cosine     & Cosine transformation of the hour to capture cyclical daily patterns. \\
        & day\_sine        & Sine transformation of the day to capture recurring daily trends. \\
        & day\_cosine      & Cosine transformation of the day to capture recurring daily trends. \\
        & month            & Month of the year. \\
    \hline
    \multirow{4}{*}{Spatial Features}
        & start\_longitude & Longitude of the starting location for the recorded activity. \\
        & start\_latitude  & Latitude of the starting location for the recorded activity. \\
        & distance\_downtown & Distance from the activity location to downtown (km). \\
        & distance\_airport  & Distance from the activity location to the airport (km). \\
    \hline
    \multirow{6}{*}{Workshift Cumulative Features}
        & shift\_earnings      & Total earnings accumulated during the current workshift (CAD). \\
        & shift\_orders        & Total number of orders received during the current workshift. \\
        & shift\_trip\_distance & Total distance traveled in trips during the current workshift (km). \\
        & shift\_idle\_distance & Total distance traveled while idle during the current workshift (km). \\
        & shift\_trip\_duration & Total duration spent on trips during the current workshift (mins). \\
        & shift\_idle\_duration & Total duration spent idle during the current workshift (mins). \\
    \hline

    \multirow{4}{*}{Weather Features}
    & temperature       & Recorded air temperature at the time of the record (°C). \\
    & precipitation     & Recorded precipitation amount during the hour (mm). \\
    & snowfall          & Recorded snowfall amount during the hour (mm). \\
    & snow\_depth       & Recorded snow‐on‐ground depth (m). \\

    \hline
  \end{tabular}
\end{table*}

\subsubsection{Ride-Hailing Records}
Driver activity and trip logs are collected from each of the major ride-hailing companies operating within the City of Toronto from January 2nd to March 13th, 2020. After deduplication, A total of 2,609  vehicles and 498,691 idle records have been collected. Each record contains a timestamped event (idle start, trip start, trip end, or log-off) associated with an anonymized vehicle identifier. By merging data across all ridehailing platforms, this study prevents driver app-switching from being misclassified as a voluntarily platform exit. An error of that kind would have occurred if the data from only one app were used.

\subsubsection{Work-Shift Cumulative Metrics}
From the chronological sequence of the idle and trip events of each driver, this study derives aggregate variables per shift that reflect the driver's current state, including the total earnings of the journey, the total duration of the trip and the total duration of the idle. These covariates describe the driver’s state and capture within subject correlation across repeated idle periods in our survival analysis.

\subsubsection{Weather Data}
Hourly weather information was gathered through the Visual Crossing Weather API, including temperature, precipitation, snowfall, and snow depth. Each ride-hailing event is matched with its nearest weather station based on location and time to provide local weather data.

\subsubsection{Spatio-Temporal Variables}
The road network and geographic data were obtained from OpenStreetMap. Spatial features are computed as the distance from each event’s origin to downtown Toronto and to Pearson International Airport. For the cyclical temporal features hour of day and day of week, sine and cosine transformations are applied to the time series data in order to maintain the integrity of the cyclical pattern.

\subsection{Study Area}
\begin{figure}[htbp]
  \centering
  \includegraphics[width=\columnwidth]{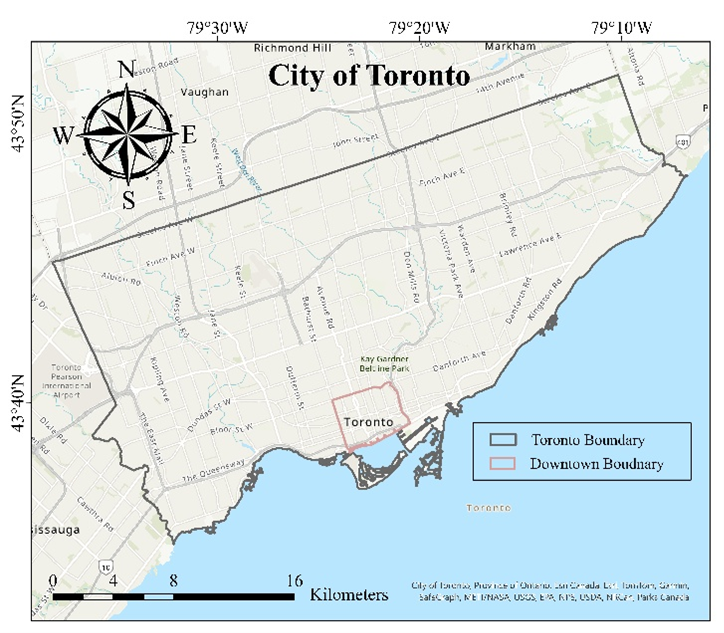}
  \caption{Study Area}
  \label{fig}
\end{figure}

Fig.~\ref{fig} illustrates the municipal boundary and major landmarks (e.g., downtown core, Pearson International Airport) used in our spatial analysis. The study area comprises the urban boundary of the City of Toronto, Canada, covering approximately 630~km$^2$ with a population of 2.9 million in 2020. Toronto features a dense road network with over 5,000 km of public streets and is served by multiple ride-hailing platforms.

\subsection{Data Processing}
All continuous features (i.e., spatial coordinates, work‐shift aggregates, and weather measurements) were standardized to remove differences in scale and units. For each feature \(x\), we computed the empirical mean \(\mu_x\) and standard deviation \(\sigma_x\) on the training set, and then transformed each observation \(x_i\) as:
\begin{equation}
\
\tilde{x}_i \;=\; \frac{x_i - \mu_x}{\sigma_x}.
\
\end{equation}

To capture the inherently cyclical nature of time‐of‐day and day‐of‐week effects without introducing discontinuities at midnight or at week boundaries, for any cyclical feature \(v\) with known period \(P\) we apply:

\begin{equation}
\begin{split}
v_{\mathrm{sin}} &= \sin\left( \frac{2\pi v}{P} \right), \\
v_{\mathrm{cos}} &= \cos\left( \frac{2\pi v}{P} \right).
\end{split}
\end{equation}

\subsection{Sample Generation and Partitioning}
Dataset construction are designed to preserve temporal integrity and to maximize the number of supervised examples

For each driver \(d\), let \(\{r_1, r_2, \dots, r_N\}\) denote their activity records sorted by increasing timestamp. A fixed-length window of size \(L\) was then applied exhaustively: for each \(i = 1, \dots, N - L + 1\), window
\begin{equation}
\
W_i = \{r_i, r_{i+1}, \dots, r_{i+L-1}\}
\
\end{equation}

is extracted. Each \(W_i\) constitutes one sample, in which the concatenated feature vectors and historical outcomes over the \(L\) time steps serve as the predictor set, and the outcome at time step \(i+L-1\) serves as the label. Advancing the window one record at a time yields \((N - L + 1)\) overlapping samples per driver, thereby fully leveraging each driver’s temporal behavior.
Once all windows \(\{W_1, \dots, W_M\}\) (where \(M = \sum_d (N_d - L + 1)\)) are generated and sorted by their start times, the complete collection was partitioned into training, validation, and test subsets in the proportions 70\,\%, 15\,\%, and 15\,\%, respectively. Specifically, the first 70\,\% of the windows were allocated for model fitting, the next 15\,\% for hyperparameter tuning, and the remaining 15\,\% for performance evaluation. By splitting on chronologically ordered windows, this procedure guarantees that no validation or test sample contains information from any future training or validation window, thereby eliminating look-ahead bias.

\section{Results}
\subsection{Descriptive Analysis}
\begin{figure*}[htbp]
    \centering
    % 第一行
    \begin{subfigure}{0.45\textwidth}
        \centering
        \includegraphics[width=\linewidth]{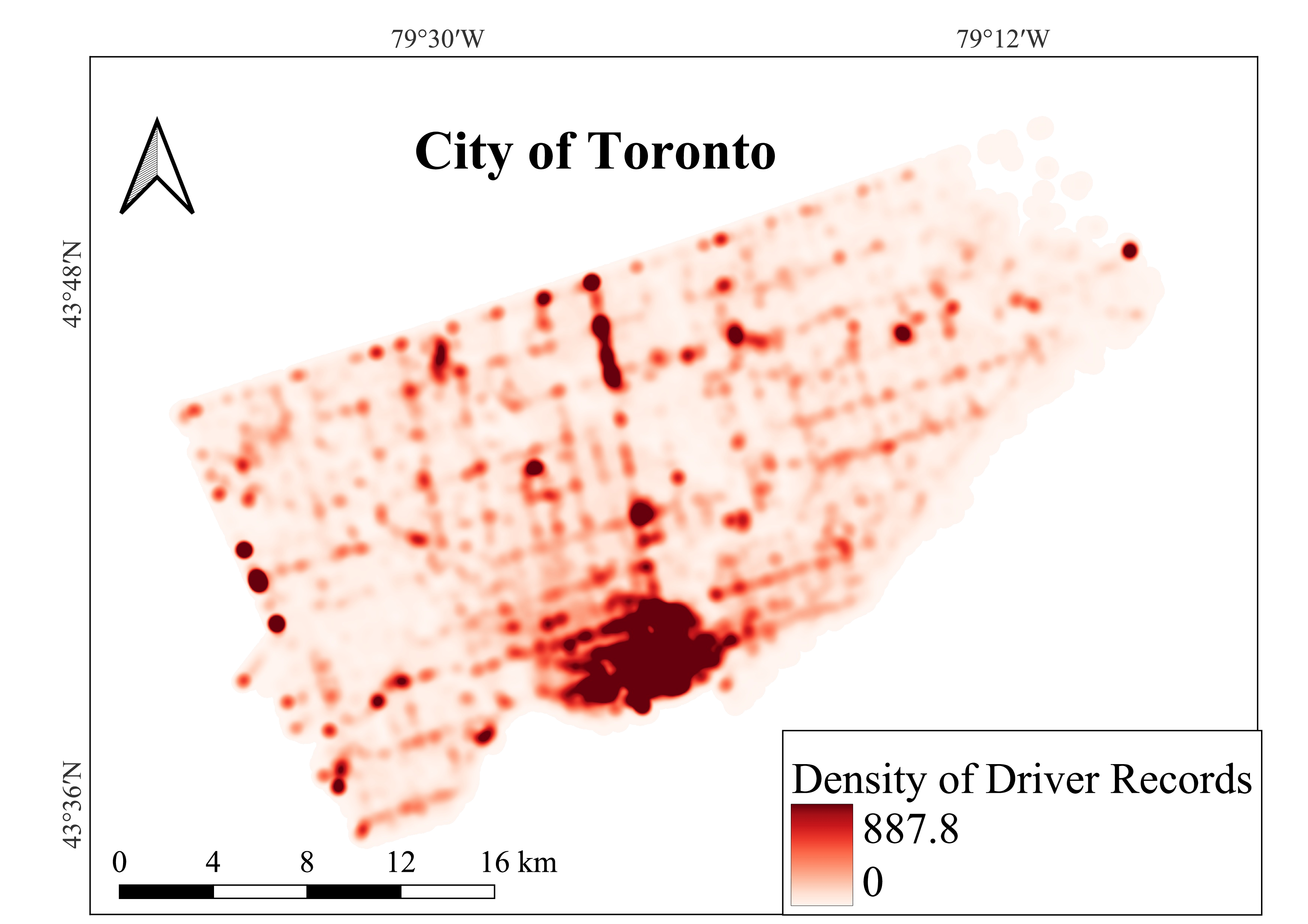}
        \caption{Spatial Density of Driver Activity Records}
        \label{fig:sub1}
    \end{subfigure}
    \hspace{2mm} 
    \begin{subfigure}{0.45\textwidth}
        \centering
        \includegraphics[width=\linewidth]{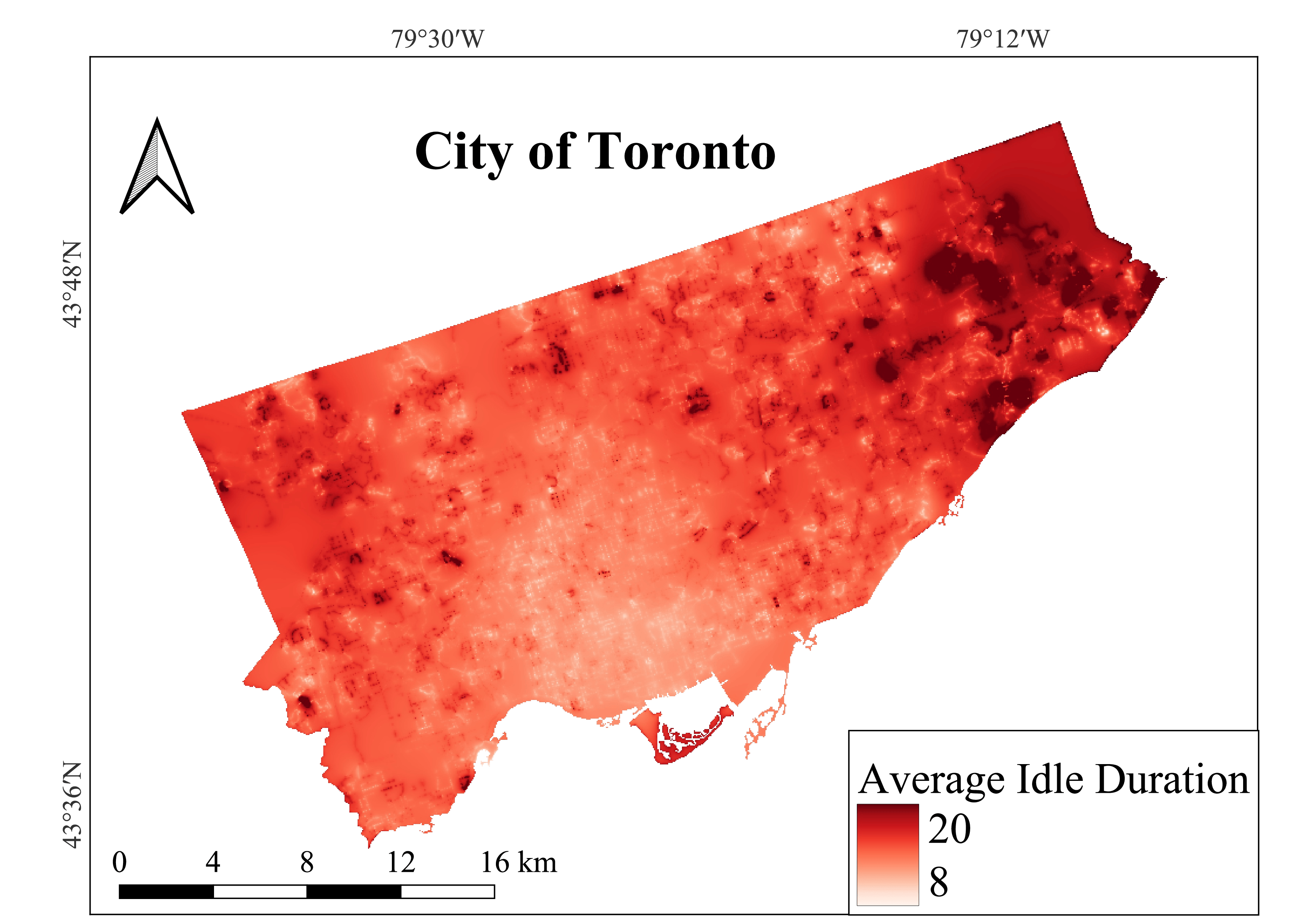}
        \caption{Average Driver Idle Duration}
        \label{fig:sub2}
    \end{subfigure}
    
    % 第二行
    \vskip\baselineskip
    \begin{subfigure}{0.45\textwidth}
        \centering
        \includegraphics[width=\linewidth]{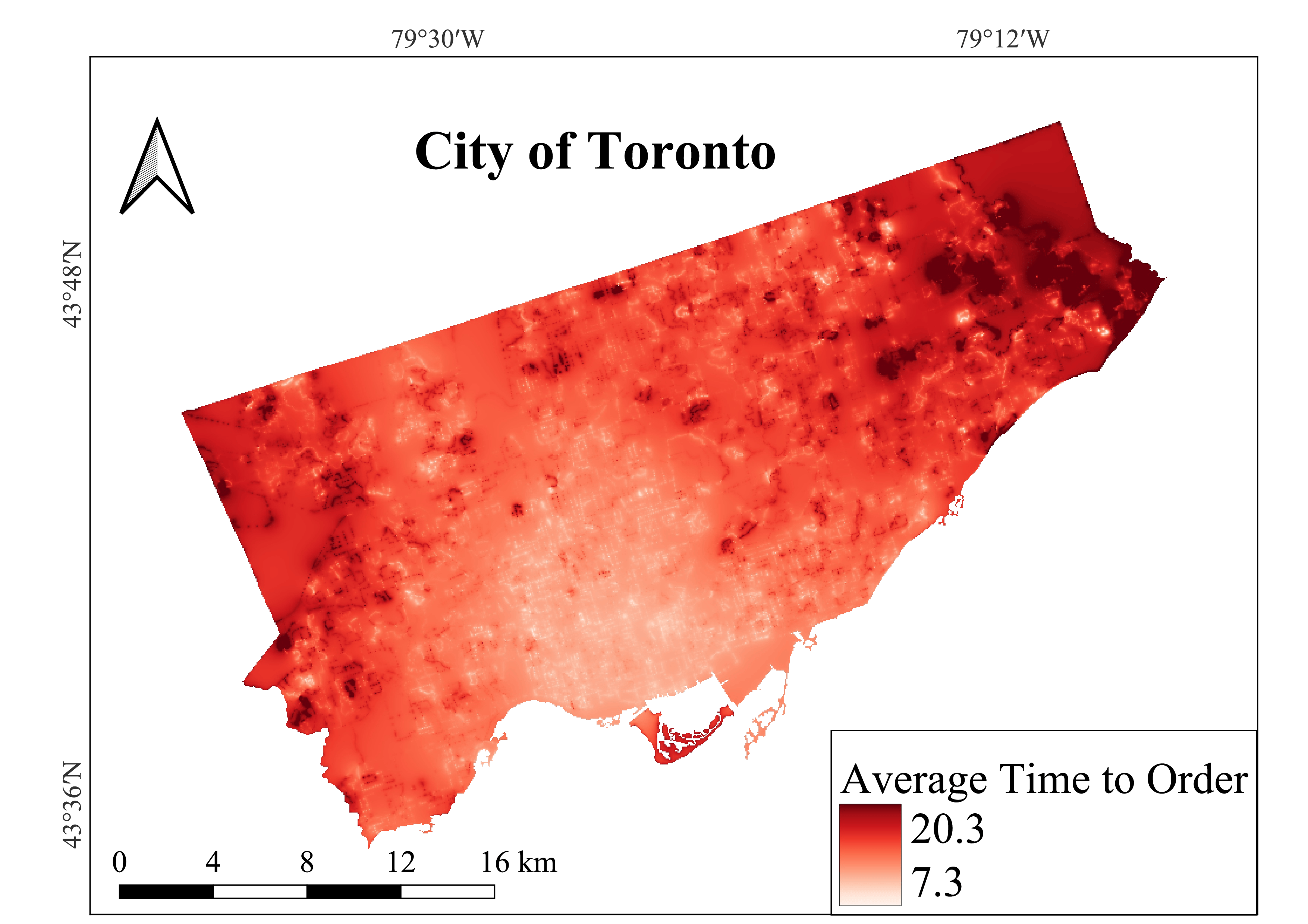}
        \caption{Average Passenger Waiting Time (Order Acceptance)}
        \label{fig:sub3}
    \end{subfigure}
    \hspace{2mm} 
    \begin{subfigure}{0.45\textwidth}
        \centering
        \includegraphics[width=\linewidth]{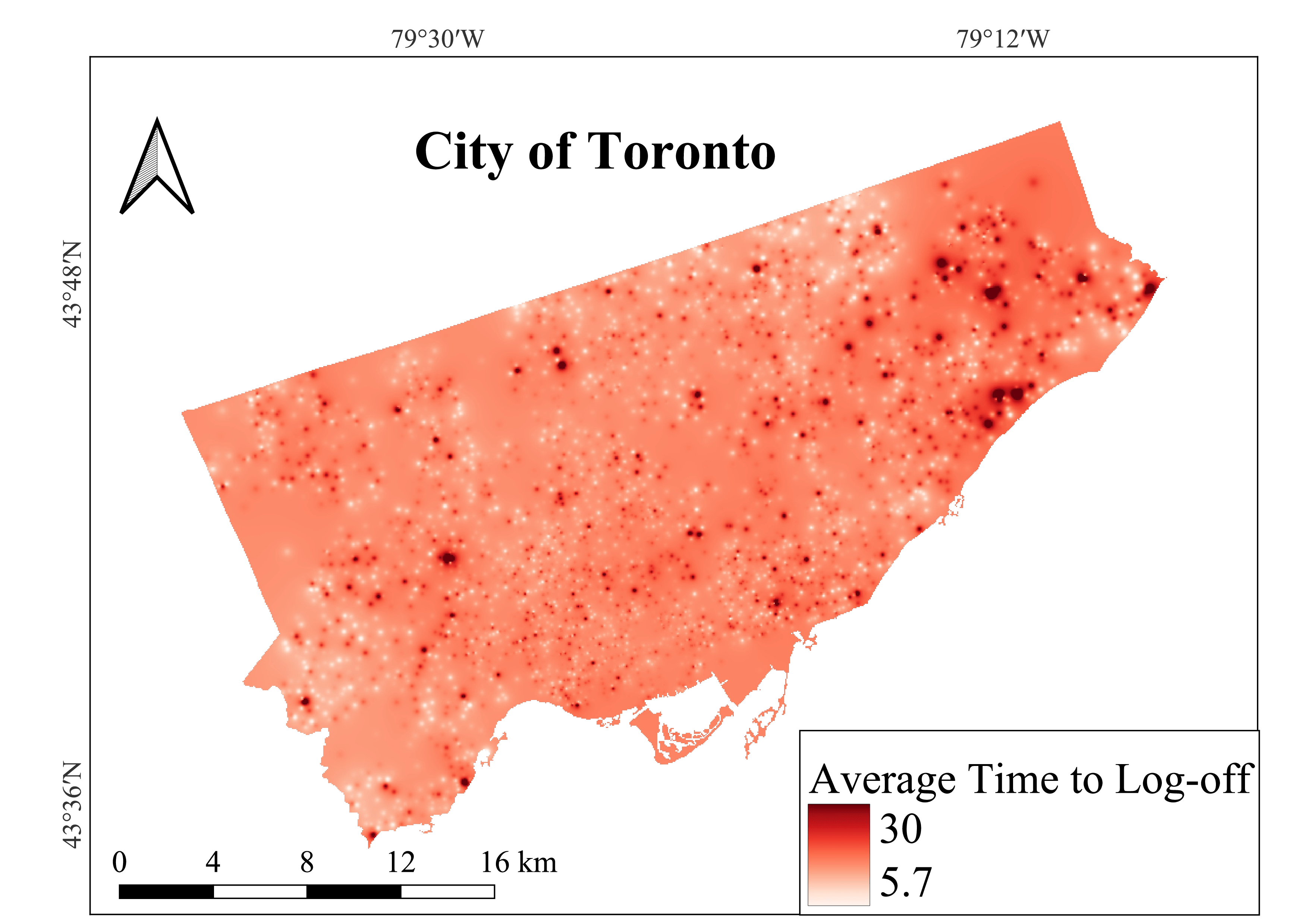}
        \caption{Average Driver Log-off Time}
        \label{fig:sub4}
    \end{subfigure}

    \caption{Spatiotemporal distribution of key supply-side indicators in Toronto’s ride-hailing system.}
    \label{fig:all}
\end{figure*}

To perform a deeper analysis into the characteristics of driver behavior, and the frequency and distribution of driver idle time events, we conducted a descriptive analysis of the dataset. 
The spatial patterns of ride-hailing driver behavior in the City of Toronto are illustrated through four geospatial heatmaps. Fig.~\ref{fig:all}\subref{fig:sub1} depicts the density of driver activity records, revealing a clear concentration in the downtown core and along major transportation corridors, indicative of higher operational intensity in central areas. Fig.~\ref{fig:all}\subref{fig:sub2} presents the average idle duration, with longer idle times observed in peripheral zones, suggesting reduced trip-matching efficiency outside the urban core. Fig.~\ref{fig:all}\subref{fig:sub3} illustrates the average time to order, which is notably lower in central regions and higher in suburban areas, reinforcing the existence of spatial demand–supply mismatches. Fig.~\ref{fig:all}\subref{fig:sub4} exhibits similar spatial patterns to those in Fig.~\ref{fig:all}\subref{fig:sub2} and \ref{fig:all}\subref{fig:sub3}, particularly between the downtown and suburban areas. This may indicate that drivers’ tolerance or patience in remaining online before logging off is relatively consistent across regions. Collectively, these visualizations highlight the spatial heterogeneity in ride-hailing system performance and underscore the operational centrality of downtown Toronto.

\begin{figure}[htbp]
  \centering
  \includegraphics[width=\columnwidth]{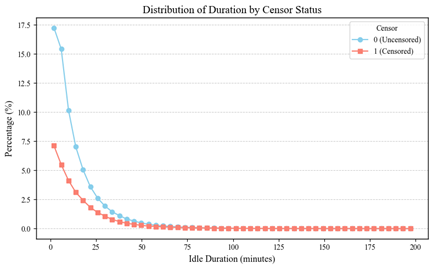}
  \caption{Illustration of censored and complete events defined in ride-hailing}
  \label{figdistribute}
\end{figure}

Fig.~\ref{figdistribute} indicates that idle behavior is far more often terminated by trip assignments than by drivers logging off. The number of active drivers declines sharply as idle time increases, with most idle sessions ending within 25 minutes through either trip acceptance or voluntary log-off. The 25-minute mark appears as a key threshold, beyond which idle sessions are uncommon and durations longer than 50 minutes are rarely observed. These patterns point to limited driver patience, suggesting a natural survival cutoff that can be incorporated into modeling assumptions.

\subsection{Kaplan-Meier Survival Estimation}
\begin{figure*}[htbp]
  \centering
  \includegraphics[scale=1]{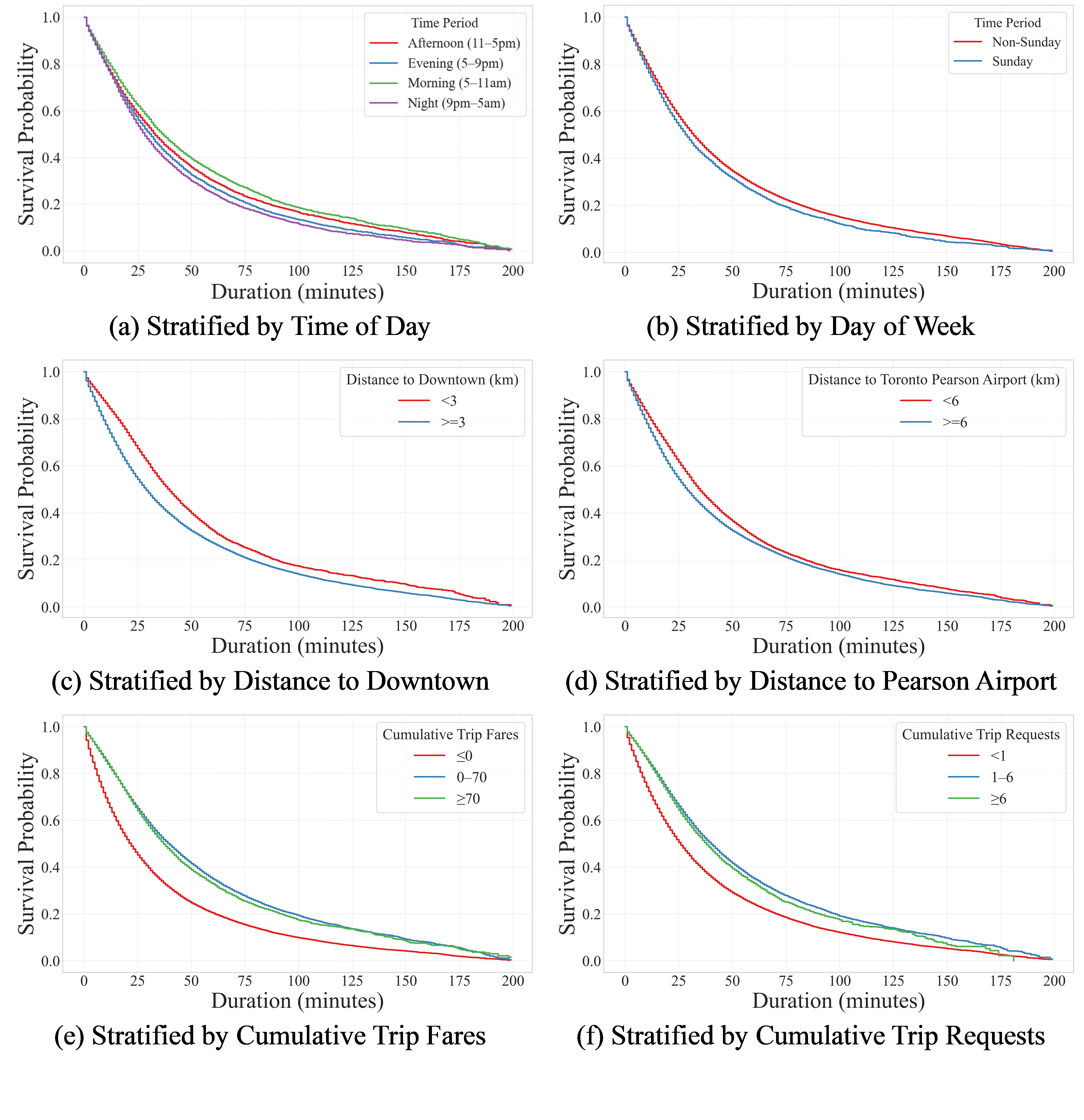}
  \caption{Kaplan–Meier Survival Curves of Trip Durations for Temporal, Spatial, and Cumulative Demand Factors}
  \label{fig:km}
\end{figure*}

The Kaplan–Meier estimates are shown in Fig.~\ref{fig:km}. The horizontal axis denotes driver session duration (minutes), and the vertical axis denotes the probability of remaining on the platform (survival probability), which declines from~1 to~0 over time. Fig.~\ref{fig:km} depicts the survival functions stratified by several factors:
\begin{itemize}
  \item \textbf{Time of day:} Four periods were defined: morning, afternoon, evening, and night. The K-M analysis suggest that the morning period exhibits the highest retention probability, followed by the afternoon, whereas the night period shows the lowest retention probability, likely owing to reduced demand and shorter trip durations.
  \item \textbf{Day of week:} Retention on Sundays is marginally lower than on other days.
  \item \textbf{Spatial zone:} The two high-demand zones of downtown Toronto and Toronto Pearson International Airport were examined. The results showed that drivers located closer to these zones were more likely to remain active on the platform.
  \item \textbf{Cumulative driver state:} Drivers who receive no trip requests during a shift have a higher probability of exiting the platform, while those who complete trips or accumulate higher fares tend to remain active. This relationship is non-monotonic: Once trip requests exceed six or cumulative fares exceed \$70,  a saturation effect emerges, which indicates a decrease in marginal incentives to stay active.
\end{itemize}
To quantify these differences, pairwise log-rank tests assuming independent samples were performed, and the results are presented in Table~\ref{tab:integral-cindex}. All group comparisons differed significantly (\(\chi^2\) test, $p<0.01$).

\begin{table}[htbp]
  \caption{All Pairwise Log–rank Test Results}
  \begin{center}
    \begin{tabular}{|c|c|c|c|}
      \hline
      \textbf{Variable}        & \textbf{Group A}           & \textbf{Group B}           & \textbf{$\chi^2$}     \\ \hline
    Trip fare              & $\le0$           & 0--70           & 13299.95 \\ \hline
    Trip fare              & $\le0$           & $\ge70$         & 7204.77  \\ \hline
    Trip fare              & 0--70          & $\ge70$         & 8.33     \\ \hline
    Trip requests          & $<1$           & 1--6            & 7891.97  \\ \hline
    Trip requests          & $<1$           & $\ge6$          & 2858.09  \\ \hline
    Trip requests          & 1--6           & $\ge6$          & 18.26    \\ \hline
    Dist.\ to DT           & $<3\,$km       & $\ge3\,$km      & 3946.39  \\ \hline
    Dist.\ to Airport      & $<6\,$km       & $\ge6\,$km      & 1337.07  \\ \hline

      Time of day              & Afternoon (11–5pm)         & Evening (5–9pm)            & 77.84           \\ \hline
      Time of day              & Afternoon (11–5pm)         & Night (9pm–5am)            & 207.33          \\ \hline
      Time of day              & Afternoon (11–5pm)         & Morning (5–11am)           & 227.70          \\ \hline
      Time of day              & Evening (5–9pm)            & Night (9pm–5am)            & 34.14           \\ \hline
      Time of day              & Evening (5–9pm)            & Morning (5–11am)           & 501.63          \\ \hline
      Time of day              & Night (9pm–5am)            & Morning (5–11am)           & 738.41          \\ \hline
      Day of week              & Non–Sunday                 & Sunday                     & 193.04          \\ \hline
    \end{tabular}
    \label{tab:logrank-summary}
  \end{center}
\end{table}

\subsection{Hyperparameters}
To determine the impact of architectural choices on model performance, a comprehensive grid search was conducted over key hyperparameters for the proposed recurrent survival model. The following components were varied: the number of attention heads \( m \in \{2, 4, 6\} \), the dimensionality of the driver embeddings \( n \in \{2, 4, 6, 8\} \), the number of Transformer encoder layers \( l \in \{1, 2, 3\} \), and the hidden size of the Transformer \(d \in \{8, 16, 32, 64\} \). Each combination was trained and evaluated using a consistent experimental setup to ensure comparability. 

The evaluation metrics was based on survival hazard ranking accuracy measured by C-index, and the results revealed that the configuration with \( m = 2 \), \( n = 4 \), \( l = 2 \), and \( d = 16 \) achieved the best overall performance.

\subsection{Effect of Sliding Window Size on Time‐Dependent Performance}
Table~\ref{tab:temporal-window-sizes} quantifies how varying the sliding window length, including no history up to previous 70 idle activities. The time-dependent discrimination (C-index) and calibration (Brier Score) have been collected for different window length at the 25th, 50th and 75th percentiles of follow-up time. 

Compared with a window size of 0 (without historical context), even a small history of previous 10 activities produces substantial gains in both C-index and Brier Score. For example,  the 25\% C index increases from 0.689 to 0.702, while the 25\% Brier score decreases from 0.096 to 0.093, demonstrating that incorporating recent driver activity records improves hazard estimation.

As the window size grows from 0 to 50, both discrimination and calibration steadily improve: the 25\% C-index rises to a peak of 0.714, and the 25\% Brier Score falls to a minimum of 0.088. 

At sizes beyond 50, the marginal gains in performance plateau, and prediction accuracy declines slightly. For example, at size 60, the 25\% C-index drops to 0.712, while the 75\% Brier score remains at 0.220.This suggests that very long windows may introduce outdated or noisy information, which can outweigh the benefit of additional history. 

Considering the diminishing returns of larger windows, we adopt a window size of 20, which preserves more samples and maintains computational efficiency by avoiding excessively long sequences.

\begin{table*}[htbp]
  \centering
  \caption{Temporal C-index and Brier Score Across Window Sizes}
  \label{tab:temporal-window-sizes}
  \small
  \setlength{\tabcolsep}{4pt} % reduce column padding
  \begin{tabular}{|c|ccc|ccc|}
    \hline
    \textbf{Window Size} & \multicolumn{3}{c|}{\textbf{C-index}}            & \multicolumn{3}{c|}{\textbf{Brier Score}}       \\
    \cline{2-7}
                         & \textbf{25\%}      & \textbf{50\%}      & \textbf{75\%}      & \textbf{25\%}     & \textbf{50\%}     & \textbf{75\%}     \\
    \hline
    0                    & $0.689 \pm 0.002$  & $0.680 \pm 0.002$  & $0.651 \pm 0.001$  & $0.096 \pm 0.000$ & $0.170 \pm 0.001$ & $0.234 \pm 0.002$ \\
    \hline
    10                   & $0.702 \pm 0.001$  & $0.694 \pm 0.000$  & $0.664 \pm 0.000$  & $0.093 \pm 0.000$ & $0.164 \pm 0.000$ & $0.227 \pm 0.001$ \\
    \hline
    20                   & $0.709 \pm 0.002$  & $0.700 \pm 0.002$  & $0.671 \pm 0.001$  & $0.091 \pm 0.000$ & $0.161 \pm 0.001$ & $0.225 \pm 0.002$ \\
    \hline
    30                   & $0.710 \pm 0.001$  & $0.702 \pm 0.001$  & $0.673 \pm 0.001$  & $0.090 \pm 0.000$ & $0.159 \pm 0.000$ & $0.222 \pm 0.002$ \\
    \hline
    40                   & $0.711 \pm 0.003$  & $0.703 \pm 0.002$  & $0.675 \pm 0.002$  & $0.089 \pm 0.000$ & $0.158 \pm 0.001$ & $0.221 \pm 0.001$ \\
    \hline
    50                   & $\mathbf{0.714 \pm 0.003}$  & $0.706 \pm 0.003$          & $\mathbf{0.677 \pm 0.002}$  & $\mathbf{0.088 \pm 0.000}$ & $0.157 \pm 0.000$ & $0.220 \pm 0.001$ \\
    \hline
    60                   & $0.712 \pm 0.003$  & $\mathbf{0.707 \pm 0.003}$ & $0.676 \pm 0.002$  & $\mathbf{0.087 \pm 0.000}$ & $0.144 \pm 0.001$ & $0.220 \pm 0.002$ \\
    \hline
    70                   & $0.712 \pm 0.001$  & $0.706 \pm 0.001$  & $0.676 \pm 0.001$  & $\mathbf{0.087 \pm 0.000}$ & $\mathbf{0.143 \pm 0.000}$ & $\mathbf{0.219 \pm 0.001}$ \\
    \hline
  \end{tabular}
\end{table*}

\begin{figure}[htbp]
  \centering
  % (a) C-index over time
  \begin{subfigure}[b]{\linewidth}
    \centering
    \includegraphics[width=\linewidth]{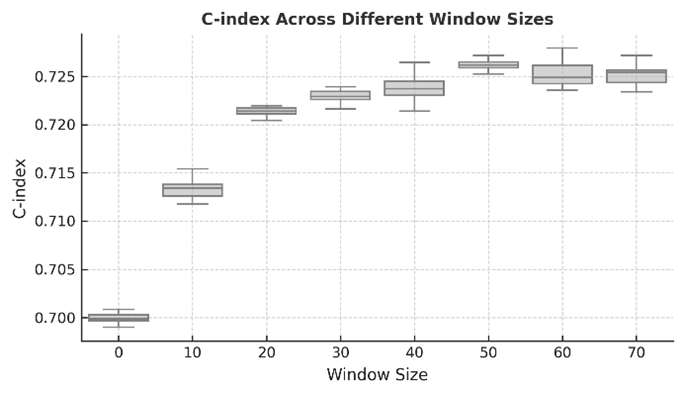}
    \subcaption{C-index over time.}
    \label{fig:cindex}
  \end{subfigure}

  \vspace{1em}  % optional separation

  % (b) Brier score over time
  \begin{subfigure}[b]{\linewidth}
    \centering
    \includegraphics[width=\linewidth]{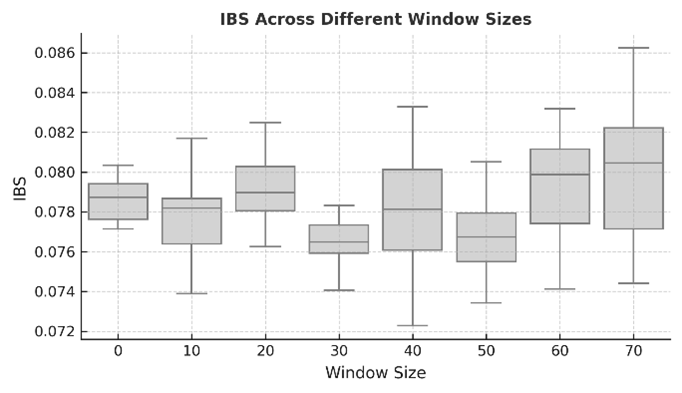}
    \subcaption{Brier score over time.}
    \label{fig:ibs}
  \end{subfigure}

  \caption{Attention weight analysis: (a) C-index, (b) Brier score.}
  \label{fig:attention-weight}
\end{figure}

\subsection{Model Comparison}
Table~\ref{tab:integral-cindex} shows the results of using the five survival models to predict driver departure hazards, using data from 20 look-back activities.

\begin{enumerate}
  \item \textbf{CoxPH:} the classical semi‐parametric Cox proportional hazards model.
  \item \textbf{DeepSurv:} a feedforward neural‐network extension of CoxPH.
  \item \textbf{LSTM-COX:} a Cox model augmented with LSTM layers to capture sequential dependencies.
  \item \textbf{Transformer-Cox:} employs multi‐head self‐attention to model long-range temporal patterns in driver activity.  
  \item \textbf{Frailty‐CoxPH:} extends the standard Cox PH with a learnable frailty term to capture latent driver heterogeneity.  
  \item \textbf{\emph{FACT} (Frailty‐Aware Cox Transformer):} combines multi‐head self‐attention for long‐range temporal modeling with driver‐specific frailty embeddings, jointly capturing temporal dynamics and latent risk factors to enhance survival prediction across recurrent events.
\end{enumerate}

Integrated C-index and Integrated Brier Score (IBS) summarize discrimination and calibration over the entire follow-up period (higher C-index and lower IBS are better). As shown in Table~\ref{tab:integral-cindex}, \emph{FACT} achieves the best overall discrimination (C-index = 0.721) and a competitive calibration (IBS = 0.080). In summary, the proposed model adequately captures key patterns found in the data, as substantiated by the C-index.
\begin{table}[htbp]
  \caption{Integral C-index and Integrated Brier Score for different models (Sliding window = 20)}
  \begin{center}
    \begin{tabular}{|c|c|c|}
      \hline
      \textbf{Method} & \textbf{C-index} & \textbf{Integrated Brier Score} \\
      \hline
      CoxPH                   & $0.607 \pm 0.001$      & $0.082 \pm 0.003$      \\
      \hline
      DeepSurv                & $0.642 \pm 0.001$      & $0.081 \pm 0.002$      \\
      \hline
    Frailty‐CoxPH                & $0.690 \pm 0.000$      & $0.081 \pm 0.002$      \\
      \hline
      LSTM-COX                & $0.707 \pm 0.002$ & $\mathbf{0.077 \pm 0.004}$ \\
      \hline
      Transformer-Cox         & $0.717 \pm 0.001$      & $0.082 \pm 0.002$      \\
      \hline
      FACT& $\mathbf{0.721 \pm 0.001}$ & $0.080 \pm 0.003$      \\
      \hline
    \end{tabular}
    \label{tab:integral-cindex}
  \end{center}
\end{table}

Table~\ref{tab:integral-cindex} shows that sequence-based Cox models generally outperform the alternatives. In particular, the frailty-based Cox PH model surpasses the traditional Cox PH model by incorporating driver-specific baseline risks. DeepSurv further improves over statistical approaches by using a neural network to capture nonlinear effects more effectively than linear regression. Notably, the Transformer-Cox model outperforms the LSTM-Cox model, which might be its self-attention mechanism that more efficiently exploits historical information.

\begin{table*}[htbp]
  \centering
  \caption{Time-dependent C-index and Brier Score (Sliding window = 20)}
  \label{tab:time-dependent}

  \begin{tabular}{|c|ccc|ccc|}
    \hline
    \textbf{Method} & \multicolumn{3}{c|}{\textbf{C-index}} & \multicolumn{3}{c|}{\textbf{Brier Score}} \\
    \cline{2-7}
    & \textbf{25\%} & \textbf{50\%} & \textbf{75\%} & \textbf{25\%} & \textbf{50\%} & \textbf{75\%} \\
    \hline
    CoxPH                 & $0.599 \pm 0.002$ & $0.590 \pm 0.001$ & $0.564 \pm 0.001$ & $0.097 \pm 0.000$ & $0.181 \pm 0.001$ & $0.260 \pm 0.003$ \\
    \hline
    DeepSurv              & $0.640 \pm 0.002$ & $0.628 \pm 0.002$ & $0.597 \pm 0.002$ & $0.096 \pm 0.000$ & $0.177 \pm 0.001$ & $0.252 \pm 0.002$ \\
    \hline

    Frailty‐CoxPH         & $0.680 \pm 0.001$ & $0.673 \pm 0.001$ & $0.645 \pm 0.001$ & $0.093 \pm 0.000$ & $0.168 \pm 0.001$ & $0.238 \pm 0.003$ \\
    \hline

    LSTM-COX              & $0.696 \pm 0.002$ & $0.685 \pm 0.002$ & $0.656 \pm 0.001$ & $0.092 \pm 0.000$ & $0.163 \pm 0.001$ & $0.229 \pm 0.002$ \\
    \hline
    Transformer-Cox       & $0.708 \pm 0.001$ & $0.698 \pm 0.001$ & $0.668 \pm 0.001$ & $\mathbf{0.091 \pm 0.000}$ & $\mathbf{0.161 \pm 0.000}$ & $0.226 \pm 0.002$ \\
    \hline
    Transformer+Embedding & $\mathbf{0.709 \pm 0.002}$ & $\mathbf{0.700 \pm 0.002}$ & $\mathbf{0.671 \pm 0.001}$ & $\mathbf{0.091 \pm 0.001}$ & $\mathbf{0.161 \pm 0.001}$ & $\mathbf{0.225 \pm 0.002}$ \\
    \hline
  \end{tabular}
\end{table*}

Because integrated metrics weight contributions from all time points equally, we further evaluate time-dependent C-index and Brier Score Brier Score at the 25\%, 50\%, and 75\% follow-up percentiles of follow-up time (Table~\ref{tab:time-dependent}). Both Transformer-based models consistently surpass the baselines at each percentile. In particular, frailty-based Cox PH model delivers the highest time-dependent C-indices (0.709, 0.700, 0.671) and the lowest Brier Scores (0.091, 0.161, 0.225) across early, median, and late follow-up. This pattern highlights a clear trend: the model predicts very accurately at short horizons but its discrimination deteriorates as the follow-up interval lengthens. As follow-up time increases, driver behavior becomes more variable and unobserved factors such as sudden log-off decisions and external disruptions play a greater role. Such external factors reduce the model's ability to differentiate high risk from low risk. Furthermore, as the duration of the follow-up interval increases, both the frequency of events and right censoring increase, which leads to a reduction in the amount of available usable information. One potential strategy to address this long-horizon performance decay may be the incorporation of time-varying covariates to capture the long term risk patterns.

\subsection{Attention Weight}
\begin{figure}[htbp]
  \centering
  \includegraphics[width=\columnwidth]{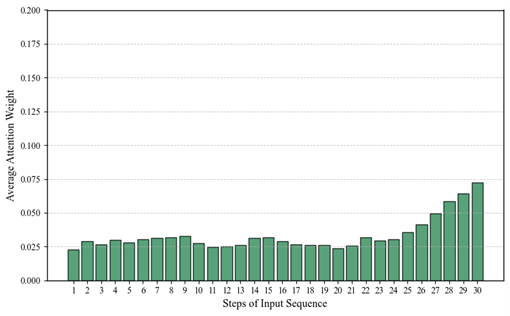}
  \caption{Illustration of censored and complete events defined in ride-hailing}
  \label{fig:attention}
\end{figure}

Average attention weights are used to analyze the influence of each activity on the predicted risk. As shown in Fig.~\ref{fig:attention}, the risk of a driver leaving the platform is primarily influenced by their most recent activities and the model assigns higher attention to recent events. 

\subsection{Ablation Study}

The ablation study was designed to assess two key aspects: first, the importance of various input feature types, and second, the effectiveness of incorporating a driver-specific frailty embedding. This embedding aims to capture latent heterogeneity and personalize baseline hazard estimates across drivers. To evaluate feature importance, we categorized the input variables into four groups, including spatial features, weather conditions, temporal encodings, and cumulative workshift metrics. Each category was individually removed, and we observed the effect on model performance. 

While prior studies have employed regression-based models such as logistic regression to estimate driver churn probability, these approaches typically lack the ability to model time-to-event dynamics and recurrent behavior. In contrast, our framework uses survival analysis, which naturally handles censored data and time-varying hazards. For this reason, we focus our comparisons on state-of-the-art survival models. 
Table~\ref{tab:ablation_cindex} indicates that all feature groups contribute meaningfully to predictive accuracy. The removal of any single group leads to a noticeable drop in performance, highlighting the complementary nature of the inputs. Furthermore, the proposed FACT model, which combines Transformer-based sequence modeling with frailty embeddings, consistently outperforms the standard Transformer-Cox baseline across all ablation scenarios. The frailty embedding proves especially beneficial in maintaining accuracy when key contextual features are omitted, underscoring its role in capturing unobserved, driver-specific risk factors.

\begin{table}[htbp]
\caption{Effect of Driver‐Specific Frailty Embedding on C‐index Across Ablation Scenarios}
\label{tab:ablation_cindex}
\centering
\footnotesize
\renewcommand{\arraystretch}{1.2}
\begin{tabular*}{\columnwidth}{@{\extracolsep{\fill}} lcc @{}}
\hline
\textbf{Ablation Scenario}      & \textbf{Without Embedding} & \textbf{With Embedding} \\
                                 & \textbf{(C-index)}     & \textbf{(C-index)}  \\
\hline
Full features                    & 0.7226                & 0.7126             \\
Temporal features                & 0.7188                & 0.7105             \\
Spatial features                 & 0.7098                & 0.7026             \\
Workshift cumulative features    & 0.7109                & 0.6964             \\
Weather features                 & 0.7220                & 0.7161             \\
No history                       & 0.7208                & 0.7044             \\
\hline
\end{tabular*}
\end{table}

\section{Conclusion}
This study explicitly formulates the behavior of drivers leaving the platform as a recurrent survival event, capturing the dynamic and flexible nature of participation in shared mobility systems. The proposed model, FACT (Frailty-Aware Cox Transformer), leverages a Transformer-based encoder with causal masking to model long-term temporal dependencies and assess how historical behavioral sequences influence the current risk of log-off. To account for individual heterogeneity, the model incorporates driver-specific frailty embeddings, which represent latent characteristics and enable more personalized risk predictions.

Our descriptive analysis revealed that drivers’ idle periods are far more often interrupted by trip assignments than by voluntary log-offs. The number of active drivers declines sharply as idle time increases, with most sessions within 25 minutes. From a platform perspective, the findings imply that dispatching strategies and retention efforts are most effective when targeted within the early stages of idleness, where both trip-matching efficiency and the risk of driver exit are most evident. Spatially, idle durations are shorter and trip‐`interruption rates are higher in high-demand areas, showing the influence of location on driver engagement.

The driver retention problem is framed through the recurrent event survival analysis, which allows the model to handle repeated log-off decisions at the individual level. To understand the influence of key factors, Kaplan–Meier analysis and log-rank tests are used to evaluate the effects of spatial attributes, temporal patterns, cumulative work status, and weather conditions. 

The Kaplan–Meier survival estimates demonstrate clear retention patterns. Morning sessions yield the highest survival probabilities, followed by afternoon and evening, with night showing the steepest drop-off. Weekly variation is modest, with Sundays exhibiting slightly lower retention. Geographically, drivers near downtown Toronto and Pearson Airport remain active longer than those in lower-demand zones. Finally, cumulative in-shift metrics matter: drivers without any trip requests exit earliest, while those completing trips or earning higher fares stay longer, additional trips or earnings yield diminishing returns. Pairwise log-rank comparisons confirm that all stratified survival curves differ significantly (\(\chi^2\) test, \(p<0.01\)). Collectively, these results highlight the influence of temporal demand patterns, geographic hotspots, and in-shift incentives on ride-hailing driver retention.

The proposed approach demonstrates strong predictive performance, with time-dependent C-index values consistently above 0.7 and low Brier Scores across early, median, and late follow-up periods. The Transformer-based encoder effectively learns behavioral patterns from historical data, and the inclusion of a learnable frailty term further enhances hazard prediction, particularly when feature availability is limited.

This model could be enhanced in many different ways. First, with the addition of real-time platform signals, such as surge pricing multipliers, supply-demand imbalances, and traffic congestion indices. It could also be changed to allow the driver’s frailty embedding to depend on time rather than be static, allowing driver characteristics to change with engagement patterns. 

Another approach to enhancing the model would be to employ randomized or quasi-experimental design to find causal relationships between platform interventions and driver retention. This might include changing incentives like targeted surge warnings, loyalty rewards, and personalized bonuses. The impact of such policy measures could then be assessed. Validation across platforms and cities, as well as between different regions and services, would also aid in model generalization and identification of regional characteristics.

Finally, to attempt to identify previously overlooked retention determinants, it might be helpful to add sociodemographic and socioeconomic data to the analysis. It might also be possible to incorporate the impact of variables like perceived algorithmic fairness, risk tolerance, and job satisfaction on operational metrics and driver logoff decisions. This data, in conjunction with a risk dashboard and real-time online learning infrastructure, could allow the platform to constantly monitor driver hazard and implement flexible interventions in response to risk.

\section*{Acknowledgment}

We appreciate the efforts of the City of Toronto in collecting and making the ride-hailing data available to our study. We greatly appreciate Matthew Lee, Raphael Dumas, and Jesse Coleman for their dedication and assistance in enhancing the quality of this manuscript.

\bibliographystyle{ieeetr}  % IEEE 风格
\bibliography{references}   % 对应 references.bib 文件名

\end{document}